\documentclass[11pt]{article}
\usepackage{coling2020}
\usepackage{times}
\usepackage{url}
\usepackage{booktabs}
\usepackage{latexsym}
\usepackage{graphicx}
\usepackage{color}
\usepackage{multirow}
\usepackage{booktabs}
\usepackage{amsmath,amssymb}
\usepackage{bm}
\usepackage{floatrow}
\usepackage{subfigure}
\usepackage{xspace} 
\usepackage{enumitem} 
\usepackage{hyperref} 
\usepackage[symbol]{footmisc}
\usepackage{pifont}
\newcommand{\cmark}{\ding{51}}
\newcommand{\xmark}{\ding{55}}

\newcommand{\ktg}{KTG\xspace}

\setlist{noitemsep,topsep=0pt}
\colingfinalcopy 

\title{Knowledge-enriched, Type-constrained and Grammar-guided Question Generation over Knowledge Bases}

\author{Sheng Bi$^{1}$ \and
Xiya Cheng$^{1}$ \and
Yuan-Fang Li$^{2}$ \and
Yongzhen Wang$^{3}$ \and
Guilin Qi$^{1*}$\\
$^{1}$School of Computer Science and Engineering, Southeast University, China \\
$^{2}$Monash University, Melbourne, Australia \\
$^{3}$Indiana University Bloomington, Bloomington, IN, USA\\
\tt\{bisheng, chengxiya\}@seu.edu.cn, \tt yuanfang.li@monash.edu,\\ \tt wang11@iu.edu, \tt gqi@seu.edu.cn}
\date{}

\begin{document}
\blfootnote{*Corresponding author.}

\maketitle
\begin{abstract}

Question generation over knowledge bases (KBQG) aims at generating natural-language questions about a subgraph, i.e.\ a set of triples. 
Two main challenges still face the current crop of encoder-decoder-based methods, especially on small subgraphs: (1) low diversity and poor fluency due to the limited information contained in the subgraphs, and (2) semantic drift due to the decoder's oblivion of the semantics of the answer entity. We propose an innovative knowledge-enriched, type-constrained and grammar-guided KBQG model, named KTG, to addresses the above challenges. In our model, the encoder is equipped with auxiliary information from the KB, and the decoder is constrained with word types during QG. Specifically, entity domain and description, as well as relation hierarchy information are considered to construct question contexts, while a conditional copy mechanism is incorporated to modulate question semantics according to current word types. Besides, a novel reward function featuring grammatical similarity is designed to improve both generative richness and syntactic correctness via reinforcement learning. Extensive experiments show that our proposed model outperforms existing methods by a significant margin on two widely-used benchmark datasets SimpleQuestion and PathQuestion. 
\end{abstract}

\section{Introduction}
\label{intro}
Question Generation over Knowledge Bases (KBQG) aims to generate natural-language questions given a subgraph in the KB, i.e.\ a set of connected triples of the form $<$subject, predicate, object$>$. 
KBQG has a wide range of applications and is increasingly attracting attention from both academia and industry. 
For example, KBQG can improve factoid-based question answering (QA) systems by either dual training of QA and QG or by data augmentation for training corpora.
As another example, KBQG can play a critical role in developing a chat-bot to ask KB-based questions under conversational settings. 

Table~\ref{example} illustrates a real scenario of KBQG, in which three questions are generated from two connected triples, along with the answer entity \textit{Ohio}. 
Among the three questions, Q3 ``Where was the high school of LeBron James, the American basketball player, located in?'' is not only correct in grammars and semantics but also more diverse than Q1
because it contains the description information of \textit{LeBron James}. Meanwhile, Q2 suffers a semantic drift problem due to a mismatch between the wrong interrogative ``when'' and \textit{Ohio} whose entity type is \textit{location}. 

\begin{table}[htbp]
\label{example}
\caption{An example of KBQG. We aim at generating questions like Q3, which is diverse and correct. Compared with Q3, Q1 has less diversity and Q2  suffers from the semantic drift problem.}
\resizebox{!}{9.5mm}{
\begin{tabular}{|c|c|c|c|}
\hline
\textbf{Input}  & \multicolumn{3}{c|}{$<$LeBron James, educated\_at, St. Vincent-St. Mary High School$> <$St. Vincent-St. Mary High School, located\_in, \textit{\textcolor{green}{Ohio}}$>$} \\ \hline
\textbf{Output} & Diverse Question        & Correct Question       & Question                                                                                     \\ \hline
Q1     &          \xmark    &        \cmark     & \textbf{Where} was LeBron James's high school located in?                                             \\ \hline
Q2     &          \cmark    &     \xmark    & \textbf{When} was the high school of LeBron James, \textit{\textbf{the American basketball player}}, located in?        \\ \hline
Q3     &          \cmark                &       \cmark                  & \textbf{Where} was the high school of LeBron James, \textit{\textbf{the American basketball player}}, located in?       \\ \hline
\end{tabular}
}
\end{table}

Recent works on neural KBQG follow an encoder-decoder architecture that takes KB subgraphs as the input to yield questions~\cite{serban2016generating}. 
In the case of entering a single triple, Elsahar et al~\shortcite{elsahar2018zero} enriched the encoder with extra contexts and equipped the decoder with attention and copy two mechanisms to improve generated questions.
For expressing given predicates and answers adequately, Liu et al~\shortcite{liu2019generating} presented a new encoder-decoder framework to incorporate diversified off-the-shelf contexts and an answer-aware loss function. 
In the case of entering multiple triples, Chen et al~\shortcite{chen2020toward} applied a bidirectional Graph2Seq model to generate questions from a KB subgraph concerning a target answer. 
While these KBQG solutions have gained noticeable success, two critical research challenges (RCs) are still under-explored to date.

\paragraph{RC-1: Limited Information.} Questions are often generated from a small KB subgraph consisting of one or a few triples, and thus the contained information may not be sufficient enough to create a diverse and fluent question well. For instance, as shown in Table ~\ref{example}, Q1 is a plain question, just a simple combination of the two connected triples, whereas Q2 and Q3 are much more informative in terms of description details. 
For this challenge, Liu et al~\shortcite{liu2019generating} expanded given triples through additional contextual information including type and range. 
However, the information inadequacy issue still matters, continuing the generation of rigid and unfluent questions. 

\paragraph{RC-2: Semantic Drift.} The semantic drift problem will occur when the semantics of generated questions becomes incompatible with given triples and/or answers. 
As shown in Table~\ref{example}, Q2 starts with the wrong interrogative ``when'' as the answer is a \emph{location} (``Ohio''). 
By contrast, Q3 starts with the correct interrogative ``where''. 
One possible reason can be that the KBQG model is trained under teacher forcing without any high-level semantic regularization.
In this manner, the resulting model may be loosely grounded by the given triples and answers. 

In this paper, we propose a novel knowledge-enriched, type-constrained, and grammar-guided KBQG model, denoted as KTG, to address the two challenges and generate correct, diverse, and fluent natural-language questions. 
For the first challenge, we augment the encoder input by linking both entities and relations in the source subgraph to an external KB, Wikidata~\cite{vrandevcic2014wikidata}, and hence introducing auxiliary knowledge such as entity \emph{description} and \emph{domain}. 
In general, the auxiliary knowledge can provide additional background information to improve the diversity and fluency of generated questions. For the second challenge, 
we label each word of a question with one of the following four types: \textbf{interrogative}, \textbf{entity word}, \textbf{relation word}, \textbf{ordinary words}, by which the decoder output is conditioned. At each decoding step, we estimate a distribution over word types first then compute multiple type-specific generation probabilities for the entire vocabulary. Meanwhile, we use a conditional copy mechanism to allow transfer content from different according to current word types. Furthermore, we conjecture that the semantics of generated questions mainly depends on the interrogative. Therefore, we explore to leverage the entity types of answers to help determine proper interrogatives. 
For instance, if the entity type of a target answer is time, then a reasonable interrogative can be ``when'' for the generated question. Besides, previous studies have utilized reinforcement learning to encourage the structural conformity between generated and ground-truth questions \cite{du2017learning,kumar2018automating}. 
This objective is achieved by promoting higher degrees of text matching evaluated typically by a \emph{rigid} reward measure such as BLEU and ROUGE. This study designs an innovative reward function based on the dependency parsing tree (DPT), which enables the proposed model to benefit from the semantic structure similarity between generated and ground-truth questions. 

The main contributions of this paper are summarized as follows.

\begin{itemize}
    \item We augment the source subgraph with auxiliary information to enrich encoder input, which improves the diversity of generated questions. 
    \item We propose to incorporate word types in generated questions, and make the decoder output conditioned on these types, which alleviates the semantic drift issue. 
    \item In a reinforcement learning framework, we design a DPT-based evaluator to encourage structural conformity whilst not rigidly enforcing subsequence matching. 
    \item We conduct extensive experiments on two benchmark datasets SimpleQuestion~\cite{bordes2015large} and PathQuestion~\cite{zhou2018interpretable} on both standard evaluation metrics and human evaluation. 
    Results demonstrate that our model outperforms state-of-the-art methods by a significant margin and that it can generate question that are more correct, diverse and fluent. 
\end{itemize}

\section{Related Work}
Our work is inspired by the recent work for KBQG based on encoder-decoder frameworks. Owing to the development of neural networks, the encoder-decoder model is initially proposed for text generation \cite{sutskever2014sequence} and has significant performances. Based on the big success of the encoder-decoder model, Serban et al.\shortcite{serban2016generating} first proposed a neural network for mapping KB fact triples into corresponding natural language questions and created the 30M Factoid Question-Answer corpus. However, their approach requires a large number of fact-question pairs as training data, which is not necessarily available for each domain. To address this challenge, Song et al.\shortcite{song2016domain} proposed an unsupervised system to generate questions from a domain-specific KB without requiring any labeled data. Besides, the types of generated questions are more diverse without any restrictions. Indurthi et al.\shortcite{indurthi2017generating} proposed an RNN based question generation model to generate natural language question-answer pairs from a knowledge graph. To generalize KBQG to unseen predicates and entity types, Elsahar et al.\shortcite{elsahar2018zero} leveraged other contexts in the natural language corpus in an encoder-decoder architecture, paired with an original part-of-speech copy action mechanism to generate questions. These contexts may make it difficult to generate questions that express the given predicate and associate with a definitive answer. Thus, Liu et al.\shortcite{liu2019generating} presented a neural encoder-decoder model that integrates diversified off-the-shelf contexts and an answer-aware loss. Finally, this model obtains significant improvements. Based on the Transformer\cite{vaswani2017attention} architecture, Kumar et al.\shortcite{kumar2019difficulty} proposed an end-to-end neural-network-based model for generating complex multi-hop and difficulty-controllable questions over knowledge graphs. Instead of using a single KB triple, Chen et al.\shortcite{chen2020toward} applied a bidirectional Graph2Seq model to generate questions from a subgraph of KB and target answers. Nevertheless, we observe that there are still two important research issues that are not processed well or even neglected, as we mentioned in sec.~\ref{intro}. Therefore, we focus on the two issues: generating diverse and fluent questions and solving semantic drift problem during the process of question generation.

Our model is also inspired by text generation from reinforcement learning (RL). RL has been successfully applied to question generation task. Pan et al.\shortcite{pan2019reinforced} proposed reinforced dynamic reasoning network, which is based on the general encoder-decoder framework but incorporates a dynamic reasoning component to generate conversational questions via an RL mechanism better. \shortcite{kumar2019putting} proposed two novel QG-specific reward functions for text conformity and answer conformity of the generated question. Besides, our work is also related to copy mechanisms. To handle rare or unknown words and copy from the input, Gu et al. \shortcite{gu2016incorporating} firstly incorporated copy mechanism into neural network-based Seq2Seq learning and propose a new model called CopyNet with encoder-decoder structure. Bao et al.\shortcite{Bao2018table} proposed KB copy to copy elements in the table (KB). Different from the above copying method, Li et al.\shortcite{li2019improving} designed a dual copy mechanism to copy from two sources with two gates to maintain the informativeness and faithfulness of generated questions.

\section{Methodology}
\begin{figure}[ht]
  \includegraphics[width=0.8\textwidth]{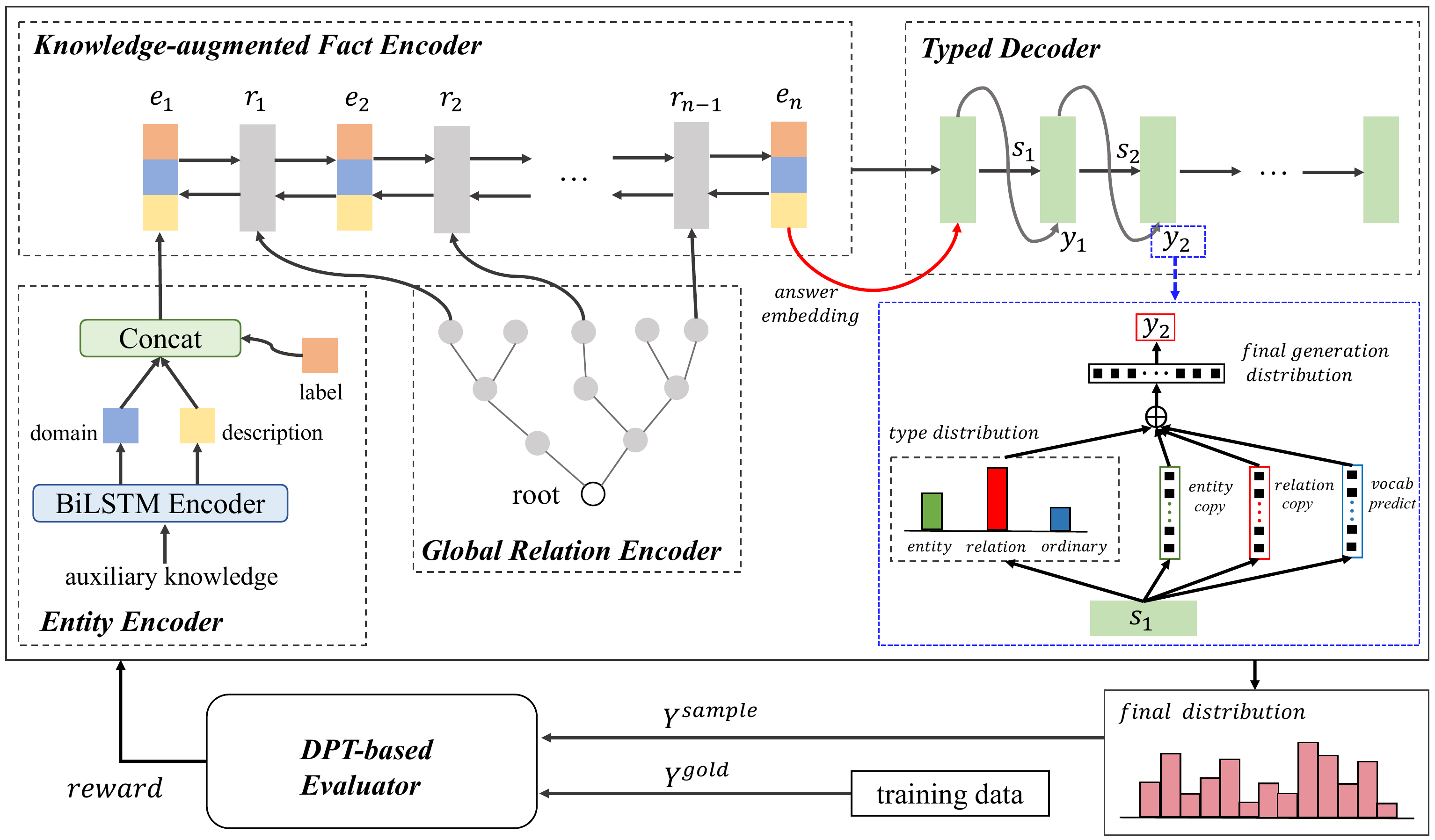}
  \caption{An illustration of our proposed model for question generation over knowledge bases.} \label{fig2}
\end{figure}

In this section, we present the details of our model. 
The overall architecture of our model is shown in Figure ~\ref{fig2}. 
Our model consists of a knowledge-augmented fact encoder, a typed decoder, as well as a grammar-guided evaluator in the reinforcement learning framework. 
The knowledge-augmented fact encoder takes the given entities, relations, and corresponding auxiliary knowledge, i.e.\ entity description and relation domain, as input and learns a knowledge-augmented fact representation. 
The learned representation is passed to a typed decoder for question generation. 
For each token the decoder outputs, the evaluator rewards the generated question using the grammatical similarity between it and the ground-truth question. 
Based on the reward assigned by the evaluator, our encoder-decoder module updates and improves its current generation.

\subsection{Problem Formulation}
In this paper, we leverage auxiliary knowledge about the input triples to generate questions over a background KB. 
We assume a collection of  triples (i.e.\ facts) $F$ as input. 
$F$ consist of two parts $E$ and $R$, where $E=\{e_1,\cdots, e_n\}$ denotes a set of entities (i.e., subjects or objects) and $R=\{r_1,\cdots, r_{n-1}\}$ denotes all the predicates (i.e.\ relations) connecting these entities. 
Moreover, $e_n\in E$ denotes the answer entity. 
Note that these facts form an answer path as a sequence of entities and relations in the KB which starts from the subject and ends with the answer: $e_1\xrightarrow{r_1}e_2\xrightarrow{r_2}\cdots\xrightarrow{r_{n-1}}e_n$. 
Given the above definitions, the task of KBQG can be formalized as follows: 
\begin{equation}
  P(Y|F)=P(Y|E, R)=\prod^{|Y|}_{t=1}P(y_t|y_{<t}, E, R, K).
\end{equation}

Here $K = (D, O)$ represents auxiliary knowledge, where $D = \left\{ \bm{\mathrm{x}^1},  \cdots, \bm{\mathrm{x}^n} \right\}$ denotes a set of entity description and $O=\left\{\bm{\mathrm{o}^1}, \cdots, \bm{\mathrm{o}^n} \right\}$ denotes the domains (i.e.\ types) for entities. $Y = (y_1, \cdots, y_{|Y|})$ is the generated question, and $y_{<t}$ denotes all previously generated question words before time-step $t$.

\subsection{Knowledge-augmented Fact Encoder}
Contrary to conventional encoders, our model takes as input not only triples but also the corresponding auxiliary knowledge as described above. 
We design a multi-level encoder to obtain the representation of knowledge-augmented facts. 
We describe our multi-level encoder below, which consists of entity encoder, relation memory, and knowledge-augmented fact encoder.

\subsubsection{Entity Encoder}\label{entity_enc}
Facts in $F$ only provide the most pertinent information of entities and relations, which is not sufficient to generate a diverse question, especially when $F$ is small. 
In this paper, we link each entity in $E$ to its respective Wikidata page, and obtain corresponding auxiliary knowledge, including a brief description, and a domain definition, to enrich the source input. 
For example, for the entity ``LeBron James'' in Table~\ref{example}, its description and domain is ``American Basketball Player'' and ``human'' respectively. 

We leverage label, description and domain information to represent each entity. 
Since the label information of an entity $e_i$ is a single token, we obtain the label embedding $\bm{\mathrm{l}_i}\in \mathbb{R}^d$ from a KB embedding matrix $\bm{\mathrm{E}_f}\in \mathbb{R}^{k \times d}$, where $k$ represents the size of KB vocabulary. 

Both description and domain information are sequences of words, and we employ a two-layer bidirectional LSTM network to encode them respectively. 
Given an entity $e_i$, its description $X^i=\left\{x_1^i, \cdots, x_{m}^i\right\}$ is a sequence of words $x_j^i$ of length $m$. 
The BiLSTM encoder calculates the hidden state at time-step $t$ by $h_t=[\overrightarrow{LSTM}([x^i_t;\overrightarrow{h}_{t-1}]);\overleftarrow{LSTM}([x^i_t;\overleftarrow{h}_{t-1}])] $.
We output the hidden state of the final time-step $h_m$ as the embedding vector, and obtain the description embedding $\bm{\mathrm{x}_i}=[\overrightarrow{h}_{m};\overleftarrow{h}_{m}]$. 
The domain embedding $\bm{\mathrm{o_i}}$ is calculated in the same way. 
The entity embedding $\bm{\mathrm{e}_i}$ is the concatenation of the label, domain and description embeddings $\bm{\mathrm{e}_i}=[\bm{\mathrm{l}_i};\bm{\mathrm{x}^i};\bm{\mathrm{o}^i}]$. 

\subsubsection{Global Relation Encoder}\label{sec:relation}
Relations in a knowledge base are typically organised hierarchically, such as $root/people/deceased\_person/place\_of\_death$. 
The global relation encoder exploits this hierarchical structure through an $N$-ary Tree-LSTM~\cite{tai2015improved} to encode these relation.
Each LSTM unit in the relation encoder is able to incorporate information from multiple child units and $N$ is the branching factor of the tree. 
Each unit (indexed by $j$) contains input and output gates $i_j$ and $o_j$, a memory cell $c_j$ and hidden state $h_j$. 
Instead of a single forget gate, the $N$-ary Tree-LSTM unit contains one forget gate $f_{jk}$ for each child $k$, $k= 1, 2,\cdots,N$, and the hidden state and memory cell of the $k$-th child are $h_{jk}$ and $c_{jk}$ respectively. 
Given the input $r_j$ in the $N$-ary Tree-LSTM, its hidden state is calculated as follows:
\begin{alignat*}{3}
  i_j&=\sigma(W^{(i)}{r_j}+\sum_{l=1}^N U_l^{(i)}h_{jl}+b^{(i)}),\quad &&f_{jk}=\sigma(W^{(f)}{r_j}+\sum_{l=1}^N U_{kl}^{(f)}h_{jl}+b^{(f)}), \\
  o_{j}&=\sigma(W^{(o)}{r_j}+\sum_{l=1}^N U_{l}^{(o)}h_{jl}+b^{(o)}),\quad &&u_{j}=tanh(W^{(u)}r_j+\sum_{l=1}^N U_{l}^{(u)}h_{jl}+b^{(u)}),\\
  c_{j}&=i_j\odot u_j + \sum_{l=1}^N f_{jl} \odot c_{jl},\quad &&h_{j}=o_j\odot tanh(c_j).
\end{alignat*}

Finally, we use the hidden state of each node $h_j$ to represent the corresponding relation embeddings $\bm{\mathrm{r_j}}$. 
In this way, the encoding is performed once and the relation embeddings are updated through backpropagation in the training process. 

\subsubsection{Knowledge-augmented Fact Encoder}
With knowledge-augmented embeddings of all entities and relations, we encode the triples $F$ using a two-layer bidirectional LSTM network with the input sequence $(\bm{\mathrm{e}_1},\bm{\mathrm{r}_1},\bm{\mathrm{e}_2},\cdots,\bm{\mathrm{r}_{n-1}},\bm{\mathrm{e}_n})$, where each $\bm{\mathrm{e}_i}$ and $\bm{\mathrm{r}_j}$ is described in Section~\ref{entity_enc} and~\ref{sec:relation} respectively. 
Note that in this paper we use a $linear$ layer to transform embeddings to maintain the consistency of embedding size. 
Ultimately, we regard the hidden states as semantic representations and obtain entity representation $(\bm{\mathrm{h}_1}, \bm{\mathrm{h}_3},\bm{\mathrm{h}_5},\dots,\bm{\mathrm{h}_{2n-1}})$ and relation representation $(\bm{\mathrm{h}_2},\bm{\mathrm{h}_4},\bm{\mathrm{h}_6},\dots,\bm{\mathrm{h}_{2n}})$. 
The last hidden state of BiLSTM is the knowledge-augmented fact representation $\bm{\mathrm{F}}$, which is fed into our decoder for question generation.

\subsection{Typed Decoder}
In order to generate questions that are consistent with the input subgraph, inspired by previous work~\cite{du2017learning}, we employ a typed decoder based on LSTM to calculate type-specific word probability distributions, which assumes that each word has a latent type of the set \{\textit{interrogative}, \textit{entity word}, \textit{relation word}, \textit{ordinary words}\}. 
In conjunction, we employ a conditional copy mechanism to allow copying from either the entity input or the relation input. 
 
At the $t$-th time-step, our decoder reads the generated word embedding $y_{t-1}$ and the hidden state $\bm{\mathrm{s}_{t-1}}$ of the previous time step to generate the current hidden state by $\bm{\mathrm{s}_t} = LSTM(\bm{\mathrm{s}_{t-1}} , y_{t-1})$. 
Note that since the first token of the generated question is interrogative, which is vital for the semantic-consistency of the generated question, we use the answer embedding, instead of the special start-of-sequence token $<$SOS$>$ embedding, at the first time step of the decoder. 
The answer embedding is the embedding of entity $e_n$, which is obtained in the entity encoder and contains label, description, and domain information. With an explicit answer embedding, the generated interrogative is more accurate, thus alleviates the semantic drift problem. 

For conditional copy from entity and relation source inputs, we leverage a gated attention mechanism to jointly attend to the entity representation and the relation representation. 
For entity representation $(\bm{\mathrm{h}_1}, \bm{\mathrm{h}_3},\bm{\mathrm{h}_5},\dots,\bm{\mathrm{h}_{2n-1}})$, the entity context vector $\bm{\mathrm{c}_t^e}$ is calculated by the attention mechanism:
  $\alpha_{t,i}^e=\frac{\exp(\bm{\mathrm{u}_t^{\top}}W_{\alpha}\bm{\mathrm{h}_i^e})}{\sum_j \exp(\bm{\mathrm{u}_t^{\top}}W_{\alpha}\bm{\mathrm{h}_j^e})}, 
  \bm{\mathrm{c}_t^e}=\sum_{i=1}^T\alpha_{t,i}^e \bm{\mathrm{h}_i^e}$,
where $W_{\alpha}$ is a trainable weight parameter. Similarly, the relation context vector $\bm{\mathrm{c}_t^r}$ can be obtained from the relation representation. Then a gating mechanism is used to control the information flow from these two sources:
\begin{equation}
  \bm{\mathrm{g}_{t}}=\sigma(W_g[\bm{\mathrm{c}_t^e};\bm{\mathrm{c}_t^r}]), \quad
  \bm{\mathrm{c}_t}=\bm{\mathrm{g}_{t}}\odot\bm{\mathrm{c}_{t}^e}+(1-\bm{\mathrm{g}_{t}})\odot \bm{\mathrm{c}_t^r}, \quad
  \bm{\mathrm{u}_t}=tanh(W_h[\bm{\mathrm{s}_t};\bm{\mathrm{c}_t}]).
\end{equation}

Generally, the predicted probability distribution over the vocabulary $V$ is calculated as: $P_V=softmax(W_V\bm{\mathrm{u}_t}+b_V)$, where $W_{V}$ and $b_{V}$ are parameters.

Different from the conventional decoder, our typed-decoder calculates type-specific generation distributions. 
Having generated the interrogative, the word types only include $\left\{ \textit{entity word}, \textit{relation word}, \textit{ordinary words} \right\}$ in the following decoding steps. 
We first estimate a type distribution over word types and decide to copy or generate words according to the word type. If the word belongs to $entity$ or $relation$, we copy this token from the input entity source or relation source. If the word is $ordinary$, we calculate type-specific generation distributions over the whole vocabulary. Finally, the generation probability is a mixture of type-specific generation/copy distributions where the coefficients are type probabilities.

We reuse the attention score $\alpha_{t,i}^e$ and $\alpha_{t,i}^r$ to derive the copy probability over entities and relations:
\begin{equation}
  P_E(\omega)=\sum_{i:\omega_{i}=\omega}\alpha_{t,i}^e, \quad
  P_R(\omega)=\sum_{i:\omega_{i}=\omega}\alpha_{t,i}^r.
\end{equation}

The final generation distribution $P(y_t|y_{<t}, F, K)$ from which a word can be sampled, is computed by:
\begin{equation}
  P(y_t|y_{<t}, E, R, K)= \sum_{g_i\in\{g_e,g_r,g_o\}} P(y_t|\tau_{y_t}=g_i, y_{<t},E, R, K)\cdot P(\tau_{y_t}=g_i|y_{<t},E, R, K).
\end{equation}
Here $\tau_{y_t}$ is the word type at time-step $t$ and $g_i$ is a word type among the three word types $\{g_{e},g_{r},g_{o}\}$. Each word can be any of the three types, but with different probabilities given the current context.

The probability distribution over three word types is calculated by: $P(\tau_{y_t}|y_{<t},F,K)= softmax(W_0\bm{\mathrm{s}_t}+b_0)$, where $W_0\in \mathbb{R}^{3 \times d}$, and $d$ is the dimension of the hidden state.
The type-specific probability distribution is computed as:
\begin{align}
  P(y_t|\tau_{y_t}=g_{e}, y_{<t},F,K) = P_E, 
  P(y_t|\tau_{y_t}=g_{r}, y_{<t},F,K) = P_R, 
  P(y_t|\tau_{y_t}=g_{o}, y_{<t},F,K) = P_V. 
\end{align}

\subsection{Evaluator}
We employ a reinforcement learning framework to fine-tunes the parameters of the encoder-decoder module by optimizing task-specific reward functions through policy gradient in the evaluator. 
Previous works directly use the final evaluation metrics BLEU, GLEU, ROUGE-L \cite{du2017learning,kumar2019putting} as rewards. 
Kumar et al.~\shortcite{kumar2019putting} also proposed the question sentence overlap score (QSS), which is the number of common n-grams between predicted question and the source sentence, as a reward function. 
Consequently, these methods tend to reward generated questions with large n-gram overlaps with the ground-truth question or the source context, thus may result in the generation of highly similar but unvaried questions. 
Therefore, we present a new reward function that is specifically designed to improve the variety of generated questions. 

\paragraph{DPTS Reward.} Dependency Parse Tree (DPT) provides a grammatical structure for a sentence by annotating edges with dependency types. 
We propose DPTS, Dependency Parse Tree Similarity, between the generated question and the ground-truth question as our reward function. 
DPTS encourages the generation of syntactically and semantically valid questions and further improve the diversity of generated questions, as it is not defined over n-gram overlapping. 
To calculate DPTS, we leverage the ACVT (Attention Constituency Vector Tree) kernel~\cite{quan2019efficient} to efficiently calculate similarity based on the number of common substructures between two trees. 

To apply the DPTS reward, we employ the self-critical sequence training (SCST) algorithm \cite{rennie2017self}. 
At each training iteration, the model generates two output sequences: the sampled output $Y^s$, in which each word $y_t^s$ is sampled according to the likelihood  $P(y_t|y_{<t}, E, R, K)$ predicted by the generator, and the baseline output $\hat{Y}$, obtained by greedy search. $r(Y)$ denotes the DPTS reward of an output sequence $Y$, and the loss function is defined as: $L_{rl}=(r(\hat{Y})-r(Y^s))\sum_t \log P(y_t^s|y_{<t}^s, E, R, K)$.

\subsection{Inference and Optimization}
Besides the loss in the evaluator, we adopt the negative log-likelihood loss function, and apply supervision on the mixture weights of word types.
\begin{equation}
  L_{cl}=\sum_t-\log P(y_t=\hat{y}_t|y_{<t}, E, R, K),\quad L_{wl}=\sum_t-\log P(\tau_{y_t}=\tau_{\hat{y}_t}|y_{<t}, E, R, K),
\end{equation}
where $\hat{y}_t$ is the reference word and $\tau_{\hat{y}_t}$ is the reference word type at time $t$. The overall loss function is defined as: $L=L_{cl}+\alpha L_{wl}+ \beta L_{rl}$, where $\alpha$ and $\beta$ are two factors to balance the three loss terms.

\section{Experiment}
In this section we present the evaluation of our KBQG model KTG. The main experiment compares our model to a number of baseline models in two settings: automatic evaluation using standard metrics, as well as human evaluation over a number of criteria. We also conduct an ablation analysis to examine the effect of various components on model performance. 

\subsection{Datasets and Preprocessing}
We conduct experiments on two widely-used benchmark datasets: SimpleQuestion~\cite{bordes2015large} dataset and PathQuestion~\cite{zhou2018interpretable}. To obtain auxiliary knowledge, we link each entity and predicate in an input subgraph to Wikidata~\cite{vrandevcic2014wikidata}, an open knowledge base, and obtain the corresponding entity description and domain and predicate hierarchy as auxiliary knowledge. In SimpleQuestion, entities are represented by their Freebase IDs. Thus we first map these Freebase IDs to Wikidata IDs and then find auxiliary knowledge according to the Wikidata IDs. PathQuestion contains verbalized entities and predicates, which can be directly used to link auxiliary knowledge. For both SimpleQuestion and PathQuestion, we add auxiliary knowledge to questions as parenthesis. As shown in Figure~\ref{example}, the italic and bold words in each question are our auxiliary knowledge. SimpleQuestion consists of over 108,000 samples and PathQuestion consists of over 11,700 samples. We randomly select 70\% of these samples for training, 10\% for validation, and 20\% for testing.

\subsection{Experimental Settings}
The size of KB embeddings and word embeddings are both set to 300.
The hidden vector size in the BiLSTM is also set to 300. 
The Adam~\cite{kingma2015adam} optimizer is used in training, with the learning rate set to 2e-5. 
Batch size and dropout rate acre set to 64, 0.5, respectively. 
We stop the training iterations until the performance difference between two consecutive iterations is smaller than 1e-6. 

\subsection{Baseline Models}
We compare our method with the following baseline models.
\begin{description}
    \item[RNN-based:] a RNN-based question generation model to generate natural language question-answer pairs from a knowledge graph~\cite{indurthi2017generating}.
    \item[Zero-Shot:] a zero-shot KBQG model for unseen predicates and entity types~\cite{elsahar2018zero}.
    \item[Multi-hop:] an end-to-end neural network-based method for automatic generation of complex multi-hop questions over knowledge graphs~\cite{kumar2019difficulty}. 
    \item[Ans-aware:] a KBQG model via using diversified contexts and answer-aware loss~\cite{liu2019generating}.
    \item[BiGraph2Seq:] a novel bidirectional Graph2Seq model to generate natural language questions from a KB subgraph and target answers.~\cite{chen2020toward}.
    \item[\ktg{}$\oplus$reward (BLEU/ROUGE/QSS):] our model that replaces DPTS with other rewards, including BLEU, ROUGE and QSS~\cite{kumar2019putting}.
\end{description}

\subsection{Evaluation Metrics}
Following previous KBQG works, we rely on a set of well-established metrics for question generation: BLEU-4 (B-4)~\cite{bgf:bleu}, METEOR (ME.)~\cite{denkowski2014meteor} and ROUGE-L (R-L)~\cite{gls:rouge} for automatic evaluation. 

Moreover, we conduct human evaluations on 50 randomly chosen questions from the test set of each dataset. 
Two human annotators were asked to judge each question on the following three criteria on a Likert scale of 1--5, with 1 being the worst and 5 being the best. \textbf{Naturalness (Nat.)} rates the fluency and comprehensibility of the generated question. \textbf{Diversity (Div.)} indicates whether the generated question contains diverse information. \textbf{Correctness (Cor.)} measures whether the question has grammar errors. 

\subsection{Results and Discussion}
The results of all evaluations are shown in Table~\ref{tab2}. 
For automatic evaluations, our model considerably outperforms all the baselines on all evaluation metrics across both datasets. The BLEU-4 score of our full model \ktg (last row) increases by 6.93 percentage points on SimpleQuestion and 5.7 percentage points on PathQuestion compared with BiGraph2Seq, which is the strongest baseline. 
Similar values can be observed for METEOR and ROUGE too. 
It is worth noting that the results of models \ktg{}$\oplus$BLEU, \ktg{}$\oplus$ROUGE and \ktg{}$\oplus$QSS are highly similar, and they outperform the baseline models. 
Yet, our full model \ktg attains superior performance, which demonstrates the effectiveness of the DPTS reward in question generation. 
For human evaluation results, our model also consistently achieves the best performance and generates significantly more natural, diverse, and correct questions. 
Our model is observed to have the highest naturalness, diversity, and correctness scores among these baseline models. 

\begin{table*}[htp]
  \centering
  \caption{Results of automatic and human evaluations on the two benchmark datasets.}
  \label{tab2}
   \resizebox{!}{22mm}
  { 
  \renewcommand\tabcolsep{5.0pt} 
  \begin{tabular}{lcccccccccccc}
    \toprule
    Datasets & \multicolumn{6}{c}{SimpleQuestion} & \multicolumn{6}{c}{PathQuestion}  \\
     \cmidrule(lr){2-7} \cmidrule(lr){8-13} 
    Metrics
    &  B-4      & ME.     &   R-L  &  Nat.      & Div.     &   Cor.  
    &  B-4      & ME.     &   R-L  &  Nat.      & Div.     &   Cor.    \\
    \midrule
    RNN-based& 19.98 &28.43 & 46.02 & 2.3 & 1.2 & 2.0 & 25.78& 33.17 &50.78  & 2.5 & 1.6 &2.2\\
   Zero-shot & 22.71  &30.39 & 51.07 & 2.6  &1.8 & 2.3 & 29.44 & 38.12  & 56.94 & 2.9 & 1.9  &2.6 \\
   Multi-hop & 25.98 &34.14 & 56.03 & 2.8 &1.9 & 2.4 & 34.14  & 41.77 &62.12 &  3.1  & 2.2 &2.8 \\
   Ans-aware & 28.19  & 36.98 & 59.17 & 3.3  &2.3 & 2.8 & 37.44 & 43.12  & 64.78 & 3.4 & 2.5  & 3.1 \\
   BiGraph2Seq & 31.12  & 39.23 & 62.14 & 3.5  &2.4 & 3.1 & 39.88 & 46.65  & 67.15 & 3.6 & 2.7  &3.2 \\
   
   \midrule
   \ktg{}$\oplus$BLEU & 34.89  & 42.55 & 65.54 & 3.9  &3.0 & 3.4 & 42.09 & 49.77  & 69.98 & 4.1 & 3.3  &3.7 \\
   \ktg{}$\oplus$ROUGE & 34.68  & 42.04 & 64.89 & 3.8  &2.9 & 3.4 & 41.67 & 49.12  & 69.24 & 4.0 & 3.2  &3.6\\
   \ktg{}$\oplus$QSS & 35.04  &43.12 & 65.99 & 3.9  &3.0 & 3.4 & 42.85 & 50.36  & 70.44 & 4.1 & 3.3  & 3.7\\
   \textbf{\ktg}& \textbf{38.05} & \textbf{46.37} &\textbf{68.13} & \textbf{4.1} & \textbf{3.2} &\textbf{3.8} & \textbf{45.58} & \textbf{52.31} & \textbf{73.21} & \textbf{4.3} & \textbf{3.5} & \textbf{4.0} \\
    \bottomrule
  \end{tabular}
  }
\end{table*}

\subsection{Ablation Test}
We conduct an ablation test to examine the effectiveness of our model components, by removing auxiliary knowledge, typed decoder, and reinforcement learning in our model one at a time. 
We can make a number of important observations from the analysis results in Table~\ref{tab3}. 
Both auxiliary knowledge in the encoder contributes and typed decoder contribute significantly and similarly to the overall model performance,  resulting a marked fall of model performance on all metrics with their removal. 
However, some subtle nuances in their contributions can be observed from human evaluation. 

The removal of the auxiliary knowledge results in the biggest reduction in both naturalness and diversity.
This is consistent with the purpose of the component, as it is designed to equip the model with more information to generate more varied questions. 
Similarly, by replacing the typed decoder with a general decoder, we observe a larger performance drop in correctness as compared to the removal of auxiliary knowledge. 
This again validates the effectiveness of the typed decoder, as it is designed to mitigate the semantic drift problem by generating correct interrogatives. 

Finally, reinforcement learning improves naturalness, diversity and correctness. 
This is due to the fact that the DPTS-based evaluator rewards high grammatical conformity (thus improves correctness), but not at the expense of enforcing n-gram similarity (thus improves naturalness and diversity). 

 \begin{table*}[htp]
  \centering
  \caption{Ablation test by removing each main component one at a time, where  ``w$/$o knowledge'' removes auxiliary knowledge from model input, ``w$/$o type'' replaces the typed decoder by a general LSTM decoder, without classifying word types, and ``w$/$o RL'' represents our model not optimized with reinforcement learning (thus without DPTS-based reward).}
  \label{tab3}
   \resizebox{!}{11mm}
  { 
  \renewcommand\tabcolsep{5.0pt} 
  \begin{tabular}{lcccccccccccc}
    \toprule
    Datasets & \multicolumn{6}{c}{SimpleQuestion} & \multicolumn{6}{c}{PathQuestion}  \\
     \cmidrule(lr){2-7} \cmidrule(lr){8-13} 
    Metrics
    &  B-4      & ME.     &   R-L  &  Nat.      & Div.     &   Cor.  
    &  B-4      & ME.     &   R-L  &  Nat.      & Div.     &   Cor.    \\
    \midrule
    \textbf{\ktg}& \textbf{38.05} & \textbf{46.37} &\textbf{68.13} &\textbf{4.1} & \textbf{3.2} &\textbf{3.8}& \textbf{45.58} & \textbf{52.31} & \textbf{73.21} &\textbf{4.3} & \textbf{3.5} & \textbf{4.0}\\
    \midrule
    w/o knowledge & 28.01 & 36.97 & 59.79 & 3.4 & 2.3 & 3.7 & 39.24 & 45.63 & 66.38 & 3.5 &2.6 &3.9\\
    w/o type & 29.27  & 39.19 & 63.58  & 3.7 &2.7 & 3.4 &  40.78 & 47.99 & 68.74 & 4.1  & 3.1  & 3.5\\
    w/o RL & 28.21 & 38.68 & 62.97 & 3.6 & 2.6 & 3.3 & 40.37 & 47.42 & 67.99 & 3.7 &3.2 &3.4\\
    \bottomrule
  \end{tabular}
  }
\end{table*}

\subsection{Case Study}
Table \ref{tab4} lists questions generated by various models for a same subgraph, providing an intuitive illustration of how our model improves the performance of question generation. 
Compared to the two baseline models Ans-aware and BiGraph2Seq, all our model variants generate questions of much higher quality. 
Among our model variants, without auxiliary knowledge, \ktg{}-knowledge only generates a \emph{plain} question without the additional information ``American actress''. 
Without the typed decoder, model \ktg{}-type generates a wrong interrogative (``what'' instead of ``where''). 
Lastly, incorporating reinforcement learning, the full model generates a syntactically and semantically valid question. 

\begin{table}[htbp] 
  \label{tab4}
  \caption{Case study.} 
   \resizebox{!}{20mm}
  {
  \begin{tabular}{l|l} 
   \toprule
   \multirow{2}{*}{Model} & $F = $ \{(laura$\_$devon, spouse, brian$\_$kelly$\_$hell), \\
           &  \quad\quad(brian$\_$kelly$\_$hell, institution, university$\_$of$\_$michigan)\} \\
   \hline 
   Ans-aware & who laura$\_$devon's spouse is?  \\
   BiGraph2Seq & what is institution laura$\_$devon's spouse?  \\
   \hline
   \ktg{}$\oplus$QSS & where does the American actress laura$\_$devon's husband work for?\\
    w/o knowledge & where does the husband of laura$\_$devon work for? \\
    w/o type & what does the husband of American actress laura$\_$devon work for?  \\
    w/o RL & where the American actress laura$\_$devon husband work for?\\
   \textbf{\ktg}& \textbf{where does the husband of laura$\_$devon, an American actress, work for?}  \\
   \hline
   \textbf{Gold}& \textbf{where does the husband of American actress laura$\_$devon work for?} \\
   \bottomrule
  \end{tabular} 
  }
 \end{table}

\section{Conclusion}
In this paper, we tackle two crucial challenges: insufficient source input and semantic drift problem for the task of question generation over knowledge bases (KBQG). 
We enrich encoder input with auxiliary knowledge, including entity descriptions and predicate domains to improve question diversity. 
We employ a typed decoder with a conditional copy mechanism to further improve the semantic-consistency of generated questions. 
We further optimize model performance through reinforcement learning and design a novel reward function based on grammatical similarity but not n-gram overlap. 
This reward ensures the generation of syntactically and semantically valid questions while allowing more diversity and fluency. 
Experimental results on two benchmark datasets show that our model achieves significant improvements over state-of-the-art models on all automatic and human evaluation metrics. 
The source code will be released to encourage reproducibility and further research \url{https://github.com/bisheng/KTG4KBQG}.
\section*{Acknowledgement}
Research in this paper was partially supported by the National Key Research and Development Program of China under grants (2018YFC0830200, 2017YFB1002801), the Natural Science Foundation of China grants (U1736204), the Judicial Big Data Research Centre, School of Law at Southeast University.

\newpage
\bibliographystyle{coling}
\bibliography{coling2020}

\end{document}